# Predicting gait events from tibial acceleration in rearfoot running: a structured machine learning approach

**Robberechts Pieter[1\*], Derie Rud[2\*], Van den Berghe Pieter[2], Gerlo Joeri[2], De Clercq Dirk[2], Segers Veerle[2], Davis Jesse[1]**

[1] Department of Computer Science, KU Leuven,
Celestijnenlaan 200A box 2402,
3001 Heverlee, Belgium

[2] Department of Movement and Sports Sciences, Ghent University,
Watersportlaan 2,
9000 Gent, Belgium

**\* Correspondence:**
Pieter Robberechts
pieter.robberechts@cs.kuleuven.be

Rud Derie
rud.derie@ugent.be

**Declarations of interest**: None.

**Acknowledgements:** This work was supported by the H2020 Interreg EU (Nano4Sports project), the Research Foundation Flanders (FWO.3F0.2015.0048.01), the KU Leuven Research Fund (C32/17/036) and the International Society of Biomechanics (matching dissertation grant program 2019).

## Abstract

**Background:** Gait event detection of the initial contact and toe off is essential for running gait analysis, allowing the derivation of parameters such as stance time. Heuristic-based methods exist to estimate these key gait events from tibial accelerometry. However, these methods are tailored to very specific acceleration profiles, which may offer complications when dealing with larger data sets and inherent biological variability. **Research question:** Can a structured machine learning approach achieve a more accurate prediction of running gait event timings from tibial accelerometry, compared to the previously utilised heuristic approaches? **Methods:** Force-based event detection acted as the criterion measure in order to assess the accuracy, repeatability and sensitivity of the predicted gait events. 3D tibial acceleration and ground reaction force data from 93 rearfoot runners were captured. A heuristic method and two structured machine learning methods were employed to derive initial contact, toe off and stance time from tibial acceleration signals. **Results:** Both a structured perceptron model (median absolute error of stance time estimation: $10.00 \pm 8.73$ ms) and a structured recurrent neural network model (median absolute error of stance time estimation: $6.50 \pm 5.74$ ms) significantly outperformed the existing heuristic approach (median absolute error of stance time estimation: $11.25 \pm 9.52$ ms). Thus, results indicate that a structured recurrent neural network machine learning model offers the most accurate and consistent estimation of the gait events and its derived stance time during level overground running. **Significance:** The machine learning methods seem less affected by intra- and inter-subject variation within the data, allowing for accurate and efficient automated data output during rearfoot overground running. Furthermore offering possibilities for real-time monitoring and biofeedback during prolonged measurements, even outside the laboratory.

## Keywords

Running, Gait event detection, Machine learning, Structured prediction

## 1    INTRODUCTION

The running gait comprises the stance and swing phases, separated by two key events: initial contact (IC) and toe off (TO) [1] (Figure 1). Determining the timing of these events allows performing a detailed stride-by-stride analysis of a runner's gait. Moreover, many variables relevant for gait analysis are defined with respect to either one or both of these gait events. For example, stance time (ST) is defined as the time span between IC and TO of the ipsilateral foot. Therefore, accurate and consistent detection of these gait events is crucial for gait analysis [2,3].

The criterion instrument for gait event detection is the force platform [1]. This expensive device is part of an instrumented runway or treadmill [4,5], hence restricted by its measurement zone. Therefore, ambulatory gait event detection methods have been developed using accelerometers that are placed on the trunk [6–9], the foot [10–12] or the tibia [13–16]. Among these locations, the tibia is most suitable to investigate shock impact and attenuation during running [17], making accelerometers attached to the tibia effective devices for screening runners on impact intensity and injury risk during in field studies.

Various heuristic methods have been proposed for determining gait events from tibial accelerations [13–15]. For example, Mercer et al. [13] identified IC as a local minimum before the axial peak tibial acceleration and TO as the second local maximum after that axial peak. A significant flaw of these methods is that they assume that both gait events are associated with consistent acceleration features, neglecting inter-subject variation [18]. Aubol and Milner [15] tried to use resultant tibial acceleration to overcome these limitations for IC detection, but obtained a similar precision to previous work. Given the success of neural networks [19–23], a data-driven approach using machine learning may lead to better precision and consistency. Therefore, the aim of this study was to determine whether structured machine learning approaches enable a more accurate and consistent detection of gait events from 3D tibial acceleration signals during rearfoot overground running. The heuristic method proposed by Mercer et al. [13] served as the baseline for the machine learning approaches and criterion validation happened by comparing the estimated timings to those determined using a force platform. We evaluated the success of event detection, the absolute error of prediction and its variability to propose an accurate estimation method.

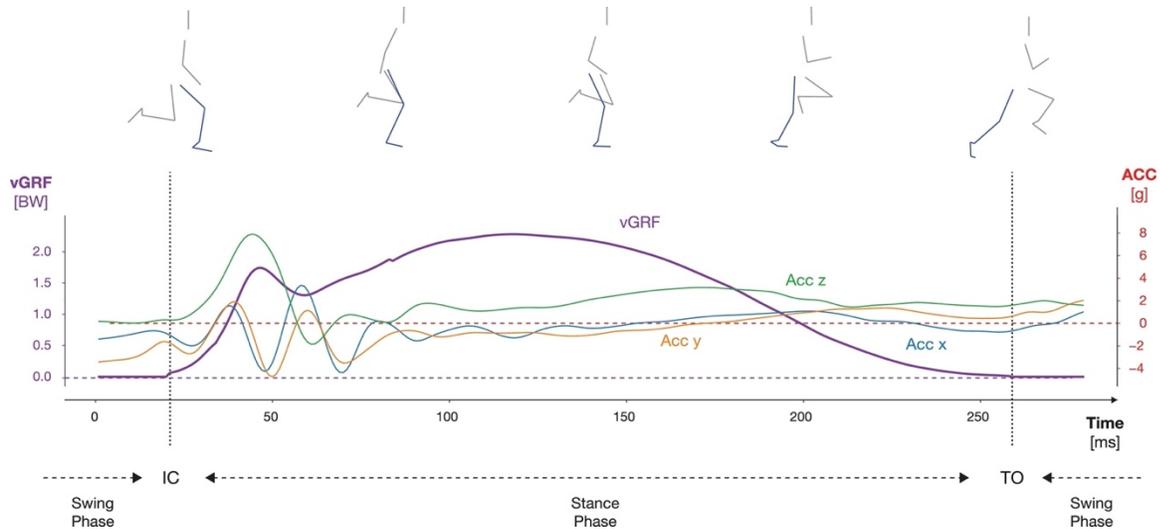

**Figure 1** The running gait cycle can be split in a stance phase and a swing phase, separated by the initial contact (IC) and toe off (TO) events. The timing of these gait events are estimated from 3D tibial acceleration signals (ACC). Ground truth labels

## 2 METHODS

### 2.1 SUBJECTS, INSTRUMENTATION AND EXPERIMENTAL PROCEDURE

This study recruited 93 rearfoot runners from the local running community (Table 1). These runners were free of running-related injuries during the last six months, ran at least 15 km per week and signed an informed consent. Approval for the study was obtained from the local ethical committee.

Table 1. Characteristics of subjects, expressed as mean ± standard deviation.

|  | Men (n = 55) | Women (n = 38) |
|---|---|---|
| Age (Yrs) | 35.9 ± 9.2 | 34.6 ± 10.8 |
| Body height (m) | 1.79 ± 0.07 | 1.67 ± 0.06 |
| Body weight (kg) | 76.5 ± 10.2 | 60.6 ± 7.3 |
| Training volume (km/week) | 36.4 ± 16.9 | 27.9 ± 11.0 |

Data collection took place during two different projects, but with the same measurement setup. The first cohort consisted of 13 subjects, wearing a standardized neutral running shoe (Li Ning Magne, ARHF041), who were asked to run on a 30-m instrumented running track at multiple speeds (2.55 m.s$^{-1}$, 3.20 m.s$^{-1}$, 5.10 m.s$^{-1}$ and preferred running speed) [5]. The second cohort consisted of 80 runners, wearing their own regular training shoes, running at 3.20 m.s$^{-1}$ only. The running speed was controlled with timing gates. Trials outside ±0.2 m.s$^{-1}$ of the target speed were rejected.

All runners wore a backpack/tablet system to measure the tibial acceleration [5]. The skin around the lower leg was pre-stretched with non-elastic tape to improve the coupling between the tibia and the sensor. Two tri-axial accelerometers (LIS331, Sparfkun, Colorado, USA;1000 Hz/axis), were tightly strapped with medical tape to the antero-medial side of both shins, eight centimeters above the medial malleolus. The axial axis was orientated along the longitudinal axis of the tibia. Simultaneously, ground reaction forces were measured at 1000 Hz by two built-in force platforms (2-m and 1.2-m, AMTI, Watertown, MA). Tibial acceleration and force data were time-synchronized by means of an infrared impulse sent from a motion capture system and captured by an infrared sensor at the backpack system [5].

## 2.2 DATA PREPROCESSING

The vertical ground reaction force was filtered using a Butterworth second-order, zero-lag, low-pass filter with a 60 Hz cutoff frequency. For each recorded step, a period ranging from 200 ms before IC to 200 ms after TO was extracted using the force data. The data from the left and right leg were mirrored. Consequently, each of these steps starts with a right foot making ground contact. This procedure resulted in 3677examples. However, for training the models, only the second step from each trial containing at least three steps was used, resulting in 1003 examples for model training. Tibial acceleration signals were filtered using a second-order band-pass filter with cutoff frequencies of 0.8 Hz and 45 Hz. Using the filter configuration as a hyper-parameter during the learning phase (Section 2.5), this configuration gave the best results.

## 2.3 FEATURE CONSTRUCTION

For each sampled value of each example, a feature vector was constructed from the x, y and z components of the bi-lateral acceleration profiles. Below we describe the features used in the final models. For the full list of considered features, we refer to the supplementary materials (A1).

**Filtered Acc {Left, Right} {x, y, z, total}**
The raw values in the filtered acceleration signals and the resultant (total) acceleration at each time step.

**Jerk {Left, Right} {x, y, z, total}**
The first derivative of the bandpass-filtered acceleration signals.

**{Roll, Pitch} {Left, Right}**
The roll and pitch extracted from the acceleration signals. Here a custom second-order Butterworth low-pass filter at 60 Hz was applied.

**Acc Right x Peak Min**
A moving average filtered labeling of local minima in the anterior-posterior x-component of the foot making ground contact. This marks the neighborhood of a clear peak value for the underlying acceleration signal.

All features were standardized by removing the mean and scaling to unit variance. This scaling happened independently on each feature and independently for each example.

## 2.4 GAIT EVENT DETECTION

Formally, the problem of gait event detection can be specified as:

> **Given:** A 3D tibial acceleration signal $x(t)$ of length $l$, described by a sequence $\bar{x} = (x_1, \ldots, x_l)$ of D-dimensional feature vectors $x_t \in \mathbb{R}^D$ with $t = 1, \ldots l$.
>
> **Find:** The gait event or phase $y_t \in$ {Swing, IC, Stance, TO} for each corresponding $x_t$, such that $\bar{y}^* = (y_1, \ldots, y_l)$ is the correct sequence of gait events and phases.

In machine learning, this type of problem is traditionally solved by multiclass classification algorithms. In this setting, the task is to find the most likely output label $y \in \{1, 2, \ldots, K\}$ for each input $x \in \mathbb{R}^D$. However, in the case of gait event detection, the output has a natural structure: IC and TO events alternate each other and the time difference between both events is similar from stride to stride. We can benefit from this inherent structure of the output to train a more accurate predictor [24].

In this "structured prediction" setting the entire output sequence of gait events and phases is predicted, rather than the label of individual samples. Specifically, given an input sequence $\bar{x}$ of a tibial acceleration signal and a corresponding possible segmentation $\bar{y} = $ (Swing, IC, Stance, …), the task is to find the element $\bar{y}^*$ of all possible output sequences $\gamma$ that maximizes a scoring function $w^T \cdot \Phi(\bar{x}, \bar{y})$:

$$\bar{y}^* = \underset{\bar{y} \in \gamma}{\text{argmax}} \ w^T \cdot \Phi(\bar{x}, \bar{y}).$$

The supplementary material (A2) describes this scoring function in greater detail.

The challenge in the structured setting is that every input $\bar{x}$ has many possible segmentations (Figure 2), making it computationally intractable to evaluate every possible input-output combination. Below, we introduce two machine learning algorithms to solve this problem.

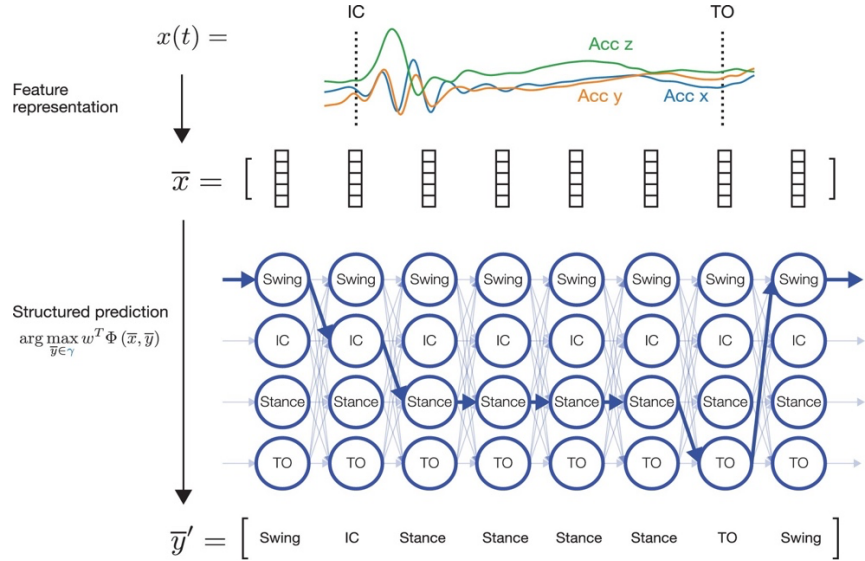

**Figure 2.** A graphical illustration of the structured prediction task. First, the acceleration signals of a step are transformed into a feature vector representation. Next, the structured prediction algorithm maps the sequence of input vectors to the most likely gait segmentation sequence. At each time step t the corresponding feature vector $\bar{x}^t$ can have any of the four possible labels. Any path through this trellis corresponds to a unique labeling of this stride. The gold standard path is drawn with bold arrows.

### 2.4.1 REFERENCE AND BASELINE METHODS

The ground truth of the IC and TO timings was determined per vertical ground reaction force. The threshold for detection was set at 20 N [14]. Additionally, we used the method defined by Mercer et al. [13] (hereafter referred to as the M-method [25]) to set a baseline for the machine learning models. This method defines clear rules for determining both the IC and TO events and performs well in comparison to other heuristic methods [25]. The IC and TO timings were herein extracted from the unfiltered vertical acceleration signal as the first local minimum before positive axial peak tibial acceleration and the minimum acceleration after a second local maximum after positive axial peak tibial acceleration, respectively. Since the vertical acceleration profile shows a fluctuating pattern with many local minima and maxima after peak acceleration in our dataset, we impose the following additional contstraints to select the most prominent minima and maxima:
- The local maxima should appear between 100 and 300 ms after peak acceleration and the distance between two maxima should be at least 100 ms.
- The local minima should appear between 20 and 200 ms after the second local maximum and the distance between two local minima should be at least 80 ms.

### 2.4.2 STRUCTURED PERCEPTRON MODEL

As a simple structured learning algorithm, we used the averaged structured perceptron algorithm [26,27] from the SeqLearn[1] package. This is a generalization of the standard perceptron algorithm and the basic building block of the deep learning algorithm introduced in the next section.

The key insight underlying the structured perceptron model is that $\Phi(\overline{x},\overline{y})$ can be decomposed as $\sum_{i=1}^{p}\Phi'_i(\overline{x},y_i)$. This makes it possible to search efficiently over all possible output sequences $\overline{y}$ using a variant of the Viterbi algorithm [28]. The joint feature functions $\Phi'_i$ are a combination of *unary features* and *Markov features* that quantify the likelihood of transitioning from one state to another in the next sample. These are learned from the data and represent, for example, that an IC event is always followed by the stance phase.

### 2.4.3 STRUCTURED RNN MODEL

Recurrent Neural Network (RNN) models [29] are a deep network architecture that can model the behavior of dynamic temporal sequences using an internal state which can be thought of as memory [30]. RNNs provide the ability to predict the current frame based on the previous and/or next frames. As such, the model can learn which long-term patterns in the acceleration profiles are relevant for determining the timings of the gait events.

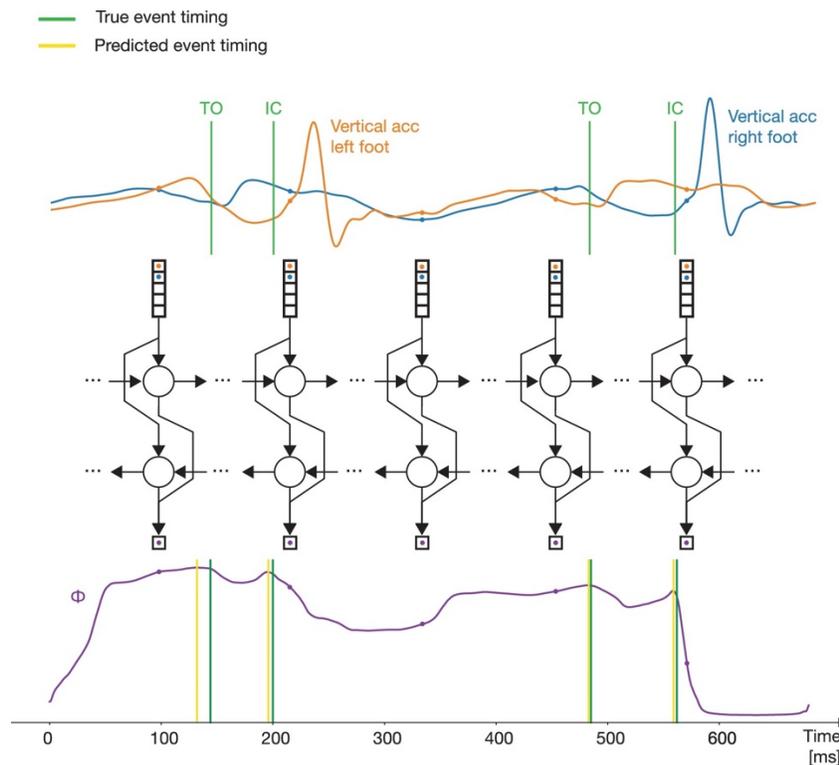

**Figure 3.** Visual representation of a two-layer bidirectional recurrent neural network with long short-term memory cells to map the multivariate time series of tibial acceleration profiles (top plot) to the likelihood of an event. Subsequently, a constrained peak detection algorithm is used to determine the most likely timing of IC and TO events. For simplicity, we only plot the axial acceleration profiles.

---

[1] http://larsmans.github.io/seqlearn/

We optimized a RNN model and a structured prediction model in an end-to-end fashion. Therefore, we use the structural hinge loss [31]

$$\bar{l}(w, \bar{x}, \bar{y}) = \max_{\bar{y}^* \in \gamma}[0, l(\bar{y}, \bar{y}^*) - w^T \Phi(\bar{x}, \bar{y}) + w^T \Phi(\bar{x}, \bar{y}^*)]$$

with

$$l(\bar{y}, \bar{y}^*) = |y_{IC} - y_{IC}^*| + |y_{TO} - y_{TO}^*|.$$

Since both the loss function and the RNN are differentiable, we can optimize them using stochastic gradient descent.

Specifically, we use the RNN outputs as feature functions for a structured prediction model (Figure 3). First, an RNN encodes the filtered acceleration signals of an entire stride and outputs a new representation for each of the samples. This new representation corresponds to an approximate likelihood of an event. Then an efficient search is executed over all possible sample values so that the most probable one can be selected. Therefore, we use a constrained peak detection algorithm which selects the local maxima that satisfy the following constraints:

- A IC and TO event of opposing foot are separated by at least 35 ms and at most 200 ms
- A TO and IC event of the same feet are separated by at least 160 ms and at most 350 ms

## 2.5 MODEL LEARNING AND PERFORMANCE EVALUATION

The machine learning models were trained and assessed in a two-step procedure. First, 5-fold cross-validation was used to obtain a good set of features and hyper-parameters for both models. For the perceptron model, the dataset was split into training (83 subjects) and test (10 subjects) sets. We found the 21 features described in section 2.3 to give the best result. The learning rate was set to 0.1, which is the only parameter of this model. For the RNN model, the dataset was randomly split into training (73 runners), validation (10 runners), and test (10 runners). The validation set was used for early stopping. The same features as in the perceptron model were used, excluding the *Acc Right x Peak Min* feature. Furthermore, we achieved the best results with AdaGrad optimization, learning rate 0.1, two bidirectional long short-term memory layers with dropout 0.2 after each recurrent layer and 50 hidden units.

Using these hyper-parameters and features, the models were retrained in a leave-one-out cross-validation analysis to evaluate the accuracy. Each model was iteratively trained on 92 of the 93 test subjects and then the accuracy of the model was tested on the 93[th] subject, obtaining an out-of-sample prediction for each subject's steps. Doing hyper-parameter tuning and feature selection for each of these 93 folds separately is not computationally feasible, hence the two-step procedure.

For each step, the relative error and absolute error were determined for the estimated IC and TO event timings [25]. Relative errors were calculated as the arithmetic difference (ms) between the predicted event timings ($t_{acc}$) obtained through the acceleration profiles and reference timings ($t_{vgrf}$) obtained through the force-platform method: Relative error $= t_{acc} - t_{vgrf}$. A positive value indicates that the detected event occurred after the reference (time lag). Absolute errors, indicating the error magnitude regardless of direction, were calculated as the absolute value of relative errors: Absolute error $= |t_{acc} - t_{vgrf}|$.

ST was determined from the estimated gait events. As for the event timings, relative and absolute errors on the estimated ST were calculated as the arithmetic difference between the estimated ST using the accelerometer-based method and reference. Here, a positive relative error corresponds to an overestimation of the ST.

The number of trials completed by each runner varied. In order to avoid that one runner would excessively impact the accuracy of our models, we computed the global median relative error and median absolute error in a two-step procedure. First, for each runner, the median absolute error and median relative error were computed over all strides of that runner. Thereafter, the global metrics were calculated as the median values of these metrics over all runners.

2.6 STATISTICAL ANALYSIS

Statistical analysis was executed in Python using the *Statsmodels* and *Scipy* libraries, with the significance level set at $p = 0.05$. A Shapiro-Wilk test for normality was first performed on the relative difference of ST. Subsequently, a Friedman test (and Wilcoxon signed-rank tests for comparing pairs of models) and Levene's test for non-normal distribution were used to examine whether the various prediction methods have significantly different accuracies and standard deviations. Post-hoc testing was conducted using Bonferroni correction. Failed predictions were imputed with the subject's average estimated ST at the corresponding running speed. Statistical analysis on the IC and TO estimates was not possible since there exists no logical imputation.

3 RESULTS

The median relative (MRE) and absolute (MAE) errors for the detected IC and TO timings and estimated ST were determined for each method (Table 2). The relative errors of ST showed a non-normal distribution (all $p <$ 0.001). Both the perceptron (MRE: 2.00 ± 11.82 ms) and RNN model (MRE: 4.00 ± 7.81 ms) outperformed the heuristic M-method (MRE: 8.00 ± 12.86 ms), with significant differences between each pair of methods (all $p <$ 0.001). Regarding the success of gait detection, the perceptron (5.11%) and RNN (3.26%) models rarely failed to identify a valid combination of IC and TO event timings. Though, more often than the M-method (0.63%). In the cases with successful identification, the M-method estimated both gait events considerably later than the criterion standard. The deviations in median relative error of the structured perceptron model and the structured RNN model are close to zero. These predicted event timings thus do not consistently lead or lag behind the true event timings.

Table 2. Median absolute error (MAE) and median relative error (MRE) ± standard deviation between each estimation method and the reference for initial contact and toe off detection (−: time lead; +: time lag) and stance time estimation (−: shorter time; +: longer time). Stance time was calculated as the timespan between the initial contact and toe off events. Additionally, the interquartile range (IQR) for the relative errors and the percentage of examples for which prediction failed are displayed.

| | | Evaluation metric | | | |
|---|---|---|---|---|---|
| Gait parameter | Method | MAE (ms) | MRE (ms) | IQR (ms) | Failed (%) |
| Initial contact (n=3677) | M-method | 8.00 ± 5.81 | 6.75 ± 8.21 | [0.50; 9.00] | 0.11 |
| | Structured perceptron | **2.00 ± 2.89** | −1.00 ± 3.59 | [-3.00; 1.00] | - |
| | Structured RNN | 2.00 ± 3.29 | −1.50 ± 3.61 | [-3.00; 0.00] | - |
| Toe-off (n=3677) | M-method | 14.00 ± 5.54 | 14.00 ± 6.85 | [12.00; 16.00] | 0.52 |
| | Structured perceptron | 9.00 ± 8.18 | 1.00 ± 10.94 | [-6.00; 4.62] | - |
| | Structured RNN | **4.00 ± 4.52** | 1.50 ± 5.87 | [-1.00; 4.75] | - |
| Stance time (n=3677) | M-method | 11.25 ± 9.52 | 8.00 ± 12.86 | [3.00; 14.00] | 0.63 |
| | Structured perceptron | 10.00 ± 8.73[*] | 2.00 ± 11.82 | [-4.50; 7.25] | 5.11 |
| | Structured RNN | **6.50 ± 5.74**[§] | 4.00 ± 7.81 | [1.00; 7.25] | 3.26 |

**Bold**: minimum MAE for detected IC and TO and estimated stance time.
[*] Significant different from M-method
[§] Significant different from structured perceptron model

Regarding the absolute error of estimation, the perceptron (MAE: 2.0 ± 2.9 ms) and RNN (MAE: 2.0 ± 3.3 ms) machine learning models better estimated the IC than the M-method (8.0 ± 5.8 ms). The TO event was harder to estimate. Still, the RNN model (MAE: 4.0 ± 4.5 ms) clearly outperformed the perceptron (MAE: 9.0 ± 8.2 ms) and M-method (MAE: 14.0 ± 5.5 ms). This RNN model also significantly outperformed the other methods in terms of error variability ($p < 0.001$). Figure 4 shows the risk of an error greater than a given threshold for the machine learning models and the heuristic M-method. In 71% of the examples, the RNN model was off by at most 10 ms whereas the perceptron model only attained this level of accuracy in 51% of the examples. Assessment of the step, stride and swing time (A3), as well as subject-specific errors (A4-6) for each method are provided in the supplementary materials.

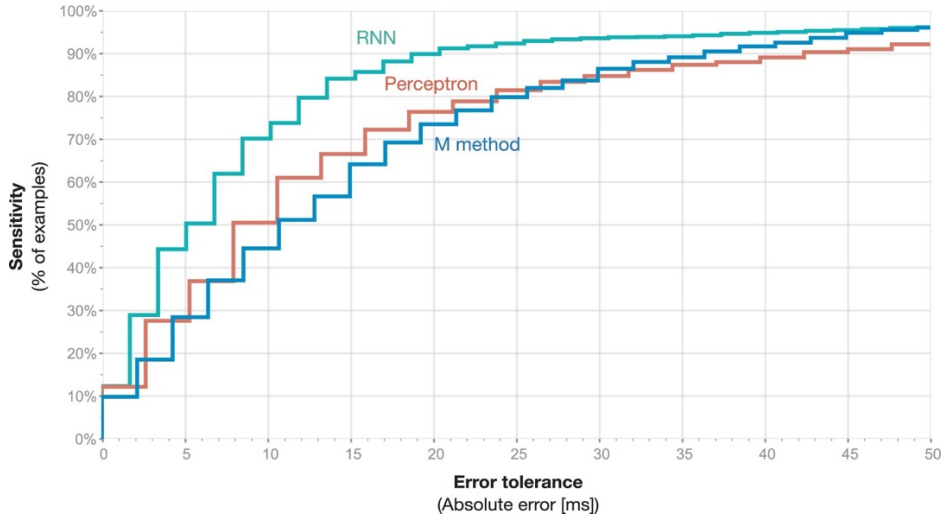

**Figure 4** Sensitivity (true positive ratio) of the three methods as a function of error tolerance. The absolute error (ms) is the deviation from the stance time, as determined by means of a force plate. Both of our machine learning models outperform the heuristic M-method.

The mean computation time of the perceptron and RNN models needed to go from the raw signals to estimated gait event timings were respectively 4 ms and 142 ms (2.3 GHz Intel Core i5, 16 GB LPDDR3 RAM).

## 4 DISCUSSION

This study examined gait event detection (IC-TO) in tibial acceleration signals of rearfoot runners. Two novel approaches for event detection were examined: a structured perceptron model with hand-crafted features and a deep learning structured RNN model. These machine learning algorithms significantly outperformed our recalculation of the heuristic M-method [13]. The visual characteristics of the signal used by the M-method cannot be generalized over a large population. Especially near TO, the acceleration profiles vary significantly between subjects, resulting in inaccurate estimates of TO, while our machine learning methods could accurately predict the gait event timings across almost all strides and subjects. This indicates that the learning methods are less affected by intra- and inter-subject variation. Although the structured learning algorithms provided accurate predictions and were more robust against variability in the tibial acceleration, the errors for an individual step could be quite large. For 3% of the examples, the RNN model predicted a ST that deviates at least 50 ms from the criterion reference (Figure 4). Most acceleration patterns of these examples belong to the same four subjects. Unfortunately, manual investigation revealed no clear patterns. One may further improve the models by adding data of a large number of runners with different, unique patterns. Alternatively, a model could be specifically trained for each pattern from a limited dataset using principles of transfer learning [32].

Given that the prediction models require data of a complete step to make a prediction and require less computation time than a typical ST in overground rearfoot running at submaximal speed, estimates can be provided before the end of the next step. This ability to output gait parameters accurately and promptly permits the development of an automated feedback system based on the consistency or fluctuation of spatio-temporal parameters. Altogether, this study presents automated methods that enable accurate and real-time detection of key events for rearfoot runners when running indoor on a sports floor. The structured RNN learning approach was most accurate compared to the other methods and offers possibilities towards implementation in overground gait analysis or gait-retraining of rearfoot runners using a durable, low-cost sensor.

This study has several limitations. Firstly, since the comparison to one heuristic method was made, one should be cautious when extrapolating these results to other existing heuristic methods. Visual inspection revealed that the acceleration profiles of the majority of our subjects did not contain the characteristics used by more advanced heuristic methods [14] making it impossible to implement a direct comparison from other methods. Secondly, all participants were rearfoot runners running on an indoor running track. Further research should investigate our proposed method when applied outdoors on different terrains and when using different footstrike patterns.

## 5  Conflict of interest statement

The authors have no conflict of interest to declare.

## 6  Acknowledgements

This work was supported by the H2020 Interreg EU (Nano4Sports project), the Research Foundation Flanders (FWO.3F0.2015.0048.01), the KU Leuven Research Fund (C32/17/036) and the International Society of Biomechanics (matching dissertation grant program 2019).

## 7  Appendix A

Supplementary material related to this article can be found, in the published version, at doi:10.1016/j.gaitpost.2020.10.035